\documentclass[11pt]{article}

\usepackage[final]{acl}
\usepackage{natbib}

\usepackage{times}
\usepackage{latexsym}
\usepackage{booktabs}
\usepackage{multicol}
\usepackage{multirow}
\usepackage{siunitx}
\usepackage{amssymb}
\usepackage[T1]{fontenc}

\usepackage[utf8]{inputenc}

\usepackage{microtype}

\usepackage{inconsolata}

\usepackage{graphicx}

\title{EvoMD-LLM: Learning the Language of Species Evolution in Reactive Molecular Dynamics}

\author{
  Zhichen Tang$^{1}$, Zhengzheng Dang$^{1}$, Yulin Chen$^{1}$, Jixin Wu$^{1}$, Haiwen Li$^{1}$, Yanming Wang\thanks{Corresponding author.}$^{2}$  \\
  $^{1}$Global College, Shanghai Jiao Tong University, Shanghai, China \\
  $^{2}$Global Institute of Future Technology, Shanghai Jiao Tong University, Shanghai, China \\
  \texttt{\{tzc233, dangzzsjtu, chenyulin, jixin\_wu, lihaiwen, yanming.wang\}@sjtu.edu.cn}
}

\begin{document}
\maketitle
\begin{abstract}
While large language models (LLMs) excel at static scientific reasoning, they struggle to model the temporal structure of dynamic physical processes. We present \textbf{EvoMD-LLM} (Evolutionary Molecular Dynamics Large Language Model), a framework that reformulates species-level molecular dynamics as a symbolic temporal language modeling problem. Reactive MD trajectories are discretized into sequences of molecular events, where each token represents a chemical species augmented with its persistence duration, enabling standard autoregressive LLMs to learn compositional evolution over time through efficient fine-tuning. A key component of EvoMD-LLM is temporal scaffolding, which treats event duration as an explicit linguistic token and serves as a structured inductive bias, significantly reducing invalid or hallucinated molecular outputs compared to conventional sequence modeling approaches. We evaluate EvoMD-LLM on multiple temporal prediction tasks, achieving up to 66.14\% accuracy and consistently outperforming sequential neural networks and language-based baselines. Beyond quantitative improvements, we qualitatively observe that the model is capable of generating interpretations for its own predictions by incorporating relevant chemical knowledge, even though it was not explicitly supervised with paired trajectory-explanation data. These results demonstrate that symbolic temporal language modeling provides an effective framework for grounding LLMs in dynamic physical simulations.

\end{abstract}

\section{Introduction}
\begin{figure}[htbp]
\centering
\includegraphics[width=1\columnwidth]{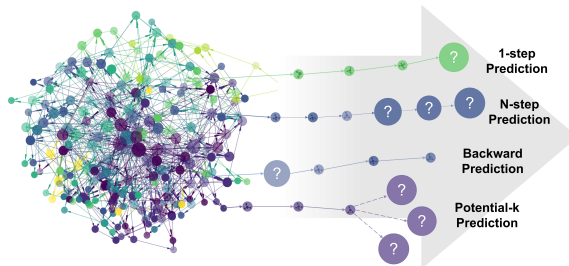}

\caption{Conceptual overview of EvoMD-LLM. The framework interprets MD trajectories as structured sequences (Nodes: species; Edges: transformations) to reconstruct reaction pathways via four predictive tasks.}
\label{fig1}
\end{figure}
The convergence of large language models (LLMs) and molecular representations has emerged as a promising direction in AI for Science. Recent paradigms have successfully aligned static molecular encodings, such as SMILES strings \cite{cavanagh2024smileyllama}, with natural language, enabling LLMs to support tasks ranging from molecular property prediction \cite{chithrananda2020chemberta} to retrieval-augmented chemical reasoning \cite{chen2025scholarchemqa}. However, most existing approaches operate on static molecular representations or rely on external tools for reasoning \cite{boiko2023emergent}. This limits their applicability to physical systems, which evolve over time through sequences of creation, persistence, and transformation events. As a result, enabling LLMs to model temporal physical processes remains a fundamental challenge in AI for Science \cite{wigh2022review}. Molecular dynamics (MD) simulations provide a natural description of temporal physical evolution by recording time-resolved atomic motions \cite{alder1957phase}. Yet, raw MD trajectories consist of high-frequency continuous coordinates that are incompatible with the discrete, symbolic token space of language models. Emerging time-series foundation models \cite{ansari2024chronos} remain inapplicable to this challenge, as their numerical quantization schemes destroy the compositional semantics and discrete identity intrinsic to chemical species. Directly aligning MD simulations with LLMs therefore presents a key abstraction challenge: how to represent continuous molecular evolution as symbolic sequences amenable to language modeling. Existing learning-based approaches to MD trajectories largely focus on structural dynamics in non-reactive or weakly reactive systems, such as protein folding \cite{tsai2020learning, bera2025accurate, murtada2024lmmol, murtada2025mdllm1}, and are ill-suited for reactive processes characterized by discrete changes in chemical species.

To address this gap, we introduce \textbf{EvoMD-LLM} (\textbf{Evo}lutionary \textbf{M}olecular \textbf{D}ynamics \textbf{L}arge \textbf{L}anguage \textbf{M}odel), a framework that reformulates species-level molecular dynamics as a constrained generative language task. We propose a modality alignment scheme that translates continuous trajectories into discrete tokens, where duration serves as an explicit semantic modifier for each chemical species. This representation enables standard autoregressive LLMs to internalize the "grammar" of chemical evolution directly through fine-tuning, eliminating the need for external simulators or specialized architectures.


A key component of EvoMD-LLM is temporal scaffolding, which explicitly encodes event duration as a linguistic token. While duration encoding is established in domains like music and speech \cite{huang2018music, ren2019fastspeech}, EvoMD-LLM introduces a fundamental shift by reframing temporal tokens as semantic proxies for kinetic stability. This structured inductive bias enables the model to internalize the underlying reaction grammar and suppress physically invalid transitions, which can be viewed as a form of semantic compression of continuous trajectories,
analogous to classical Run-Length Encoding (RLE) schemes in data compression \cite{sayood2017data}. Empirical ablation studies(Section \ref{sec:ablation}) show that this design significantly improves prediction accuracy and reduces invalid or hallucinated molecular outputs.

We evaluate EvoMD-LLM on a comprehensive suite of temporal prediction tasks, as illustrated in Figure~\ref{fig1}. Beyond quantitative metrics, we remarkably observe that the model exhibits emergent explanatory behaviors: despite lacking explicit supervision, it spontaneously produces plausible physical rationales for kinetic stability. These results demonstrate that symbolic temporal language modeling serves as an effective framework for learning species-level dynamics. Our main contributions are summarized as follows:
\begin{itemize}
    \item \textbf{EvoMD-LLM Framework:} We propose a language modeling framework that reformulates species-level molecular dynamics as symbolic event sequences, enabling standard autoregressive large language models to model temporal evolution in reactive systems.

    \item \textbf{Temporal Scaffolding via Duration Tokens:} We introduce temporal scaffolding by explicitly encoding event persistence as linguistic tokens. This structured inductive bias significantly improves prediction accuracy and reduces invalid molecular outputs, as demonstrated by extensive ablation studies.

    \item \textbf{Unified Temporal Prediction Formulation:} We show that a single instruction-tuned language model can flexibly support diverse temporal prediction tasks, including forward forecasting and backward inference, without task-specific architectures.
\end{itemize}

\section{Methods}
\label{sec:Methods}

\begin{figure*}[htbp]
\centering
\includegraphics[width=0.97\textwidth]{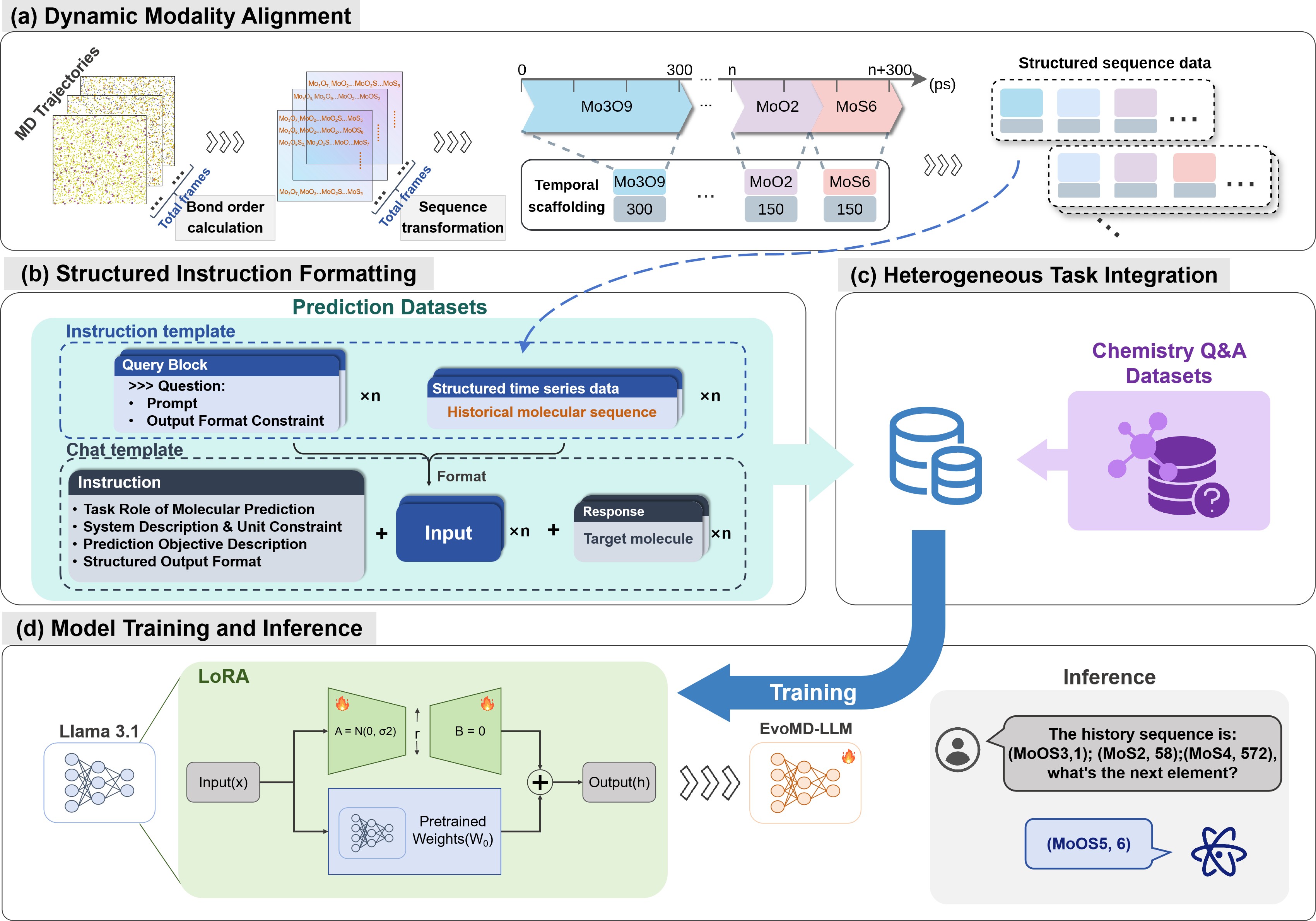}
\caption{The overall framework of the model. Encompassing dynamic modality alignment, structured instruction formatting,  heterogeneous task integration and model training along with inference.}
\label{fig2}
\end{figure*}

We propose EvoMD-LLM to treat molecular evolution as a foreign language with its own grammar of causality and persistence.
As illustrated in Figure \ref{fig2}, our framework operates through a four-stage pipeline: (1) Dynamic Modality Alignment; (2) Structured Instruction Formatting; (3) Heterogeneous Task Integration; and (4) Model Training and Inference. In this section, we detail the theoretical formulation and key algorithmic components.

\subsection{Problem Formulation}
\label{sec:formulation}

We enable LLMs to learn the dynamics of chemical reactions by reformulating MD simulations as a structured symbolic text generation problem.

A standard MD simulation produces a raw trajectory $\mathcal{T}_{\text{raw}}$, recording atomic positions $\mathbf{R}$ and momenta $\mathbf{P}$ at each time step $\tau$:
\begin{equation}
\mathcal{T}_{\text{raw}} = \{ (\mathbf{R}(\tau), \mathbf{P}(\tau)) \mid 0 \le \tau \le T \}.
\end{equation}
While physically complete, such trajectories are high-dimensional and dominated by thermal noise, which obscures long-term reaction patterns. To obtain a representation amenable to language modeling, we apply a transformation $\Phi$ that maps raw trajectories to a discrete sequence of molecular states:

\begin{equation}
    \mathcal{X} = \Phi(\mathcal{T}_{\text{raw}}) = \{ (m_i, \Delta t_i) \}_{i=1}^{N},
\end{equation}
where $N$ denotes the number of discrete events in the transformed sequence, $m_i \in \mathcal{V}$ is a molecular-formula token drawn from the chemical vocabulary $\mathcal{V}$, and $\Delta t_i \in \mathbb{Z}^{+}$ is the persistence duration of that event measured in picoseconds (ps). This abstraction suppresses high-frequency atomic fluctuations while preserving the causal sequence of chemical transformations. Unlike standard text generation where tokens are equidistant, chemical evolution is an irregularly sampled time series. We treat this sequence directly as natural language. This allows us to train the model using standard autoregressive cross-entropy loss, without requiring specialized regression architectures.

\paragraph{Generative Modeling Objective.}
We formulate reaction modeling as conditional sequence generation. Given a context sequence $\mathbf{x}$ and an instruction $\mathcal{I}$, the model generates a target sequence $\mathbf{y}$ according to the factorization:
\begin{equation}
P(\mathbf{y} \mid \mathbf{x}, \mathcal{I}) = \prod_{j=1}^{|\mathbf{y}|} P(y_j \mid y_{<j}, \mathbf{x}, \mathcal{I}),
\end{equation}
where $\mathbf{y} = ((m'_1, \Delta t'_1), \dots, (m'_{|\mathbf{y}|}, \Delta t'_{|\mathbf{y}|}))$ represents the target sequence. The instruction $\mathcal{I}$ specifies the task (e.g., forward or backward prediction), enabling a unified formulation across different reaction reasoning scenarios.

\subsection{Dynamic Modality Alignment}
\label{sec:alignment}

To bridge the gap between continuous physical simulations and discrete symbolic reasoning, as illustrated in figure~\ref{fig2}(a), we construct a Dynamic Modality Interface. This process translates raw MD trajectories into a structured "grammar" of reaction events, characterized by semantic identity and temporal persistence.

\paragraph{From Continuous Trajectories to Discrete Events.}
Raw MD data consists of high-frequency atomic coordinates dominated by thermal noise. We adopt the \textit{ab initio} bond-order determination method established by \newcite{dang2025unraveling} as the physical ground truth for identifying atomic connectivity, where the total bond order between atoms $i$ and $j$ is decomposed as $BO_{ij} = BO_{ij}^{\sigma} + BO_{ij}^{\pi} + BO_{ij}^{\delta}$. An atomic pair is treated as bonded only when $BO_{ij} > BO_{\min}$, which yields the frame-wise connectivity used for downstream species extraction. Concretely, each MD frame is converted into an undirected graph $G=(V,E)$, where atoms are nodes and valid bonds are edges; we then apply depth-first search (DFS) to identify connected components, each of which is serialized into a molecular formula. Building upon these snapshots, our framework projects the continuous evolution into a discrete event space by defining molecular formulas as atomic semantic units. Unlike standard NLP approaches that tokenize chemical strings into sub-word units (e.g., SMILES characters \cite{cavanagh2024smileyllama}), we treat each distinct molecular formula as an atomic semantic unit. This preserves the integrity of chemical identity, allowing the LLM to reason over species-level transformations rather than character-level statistics.



We define a valid Molecular Event $\mathcal{E} = (m, \Delta t)$ as a tuple comprising a molecular species $m$ and its persistence duration $\Delta t$. To distill chemically significant states from transient thermal fluctuations, we treat events with $\Delta t < \tau_{\min}$ as high-frequency noise and retain only band-pass filtered events satisfying $\tau_{\min} \leq \Delta t \leq \tau_{\max}$, with $(\tau_{\min}, \tau_{\max}) = (10, 500)$ ps. The lower cutoff removes sub-10-ps fluctuations that mainly reflect bond vibrations rather than chemically meaningful state changes, while the upper cutoff excludes overly persistent plateaus that dominate the raw trajectory and obscure the intermediate reaction dynamics of interest. This operation effectively isolates stable reaction intermediates from high-frequency noise while excluding ultra-short-lived vibrations and overly persistent plateau states. Details about the original dataset scale and filtering statistics are provided in Appendix~\ref{app:data_stats}.

\paragraph{Structured Context Construction.}

To enable autoregressive forecasting, the discrete event stream is segmented into structured input-output pairs using a sliding window approach. Each training example consists of a historical context window (3-5 events) and a target future event.

Raw reaction data exhibits a long-tail distribution: a few stable species dominate, while key transition states are rare. To avoid trivial frequency-based prediction, we apply two-dimensional stratified sampling over molecular identity and temporal regimes, where temporal regimes are discrete duration bins spanning short-, medium-, and long-lived events. Each stratum is sampled toward a more uniform count distribution before constructing the final training windows, improving coverage of both rapid intermediates and stable products.

Detailed visualizations of the data evolution, species distribution, and the effects of balancing are presented in Appendix~\ref{app:data_stats} (Figure \ref{fig3}).




\subsection{Temporal Scaffolding}
\label{sec:scaffolding}

Standard Transformers, while adept at sequence ordering, remain agnostic to variable time intervals. To bridge this gap, EvoMD-LLM implements Temporal Scaffolding by interleaving species tokens with duration tokens ($\Delta t_i$), reinterpreting this strategy as a neural implementation of Run-Length Encoding (RLE) for semantic trajectory compression \cite{sayood2017data}. While this design shares structural similarities with variable-duration modalities such as note sustain in Music Transformers \cite{huang2018music} and phoneme alignment in FastSpeech \cite{ren2019fastspeech} , EvoMD-LLM introduces a novel Kinetic-to-Semantic Mapping that treats duration as an intrinsic indicator of kinetic stability. This provides a structured inductive bias that enforces physical consistency and suppresses "kinetic hallucinations" by differentiating between thermodynamically stable states and transient intermediates. The functional necessity of this design is empirically validated by our ablation study (Section \ref{sec:ablation}), where removing duration tokens results in a sharp 11.67\% absolute decrease in 1-step accuracy (from 66.14\% to 54.47\%). This formulation effectively decouples continuous physical time from the logical reaction sequence, enabling the model to skip redundant noise and reason directly across chemically significant timescales.

\subsection{Structured Instruction Formatting}
\label{sec:prompting}

To transform the interleaved event sequences into training samples, we employ a structured instruction-tuning paradigm. As illustrated in Figure \ref{fig2}(b), we design a domain-specific template that enforces strict syntactic constraints on the generative output.

The construction consists of two components:
\begin{enumerate}
    \item \textbf{System Context (Semantic Definition):} We utilize the system prompt to define the model's role as a "Scientific Simulator." Crucially, this prompt establishes the semantic mapping for our unified vocabulary, explicitly instructing the model that the output must alternate between molecular formulas (representing state identity) and time tokens (representing kinetic stability).
    \item \textbf{Task Instruction (Historical Constraints):} The user prompt encapsulates the historical context window $x = \mathcal{H}_{<t}$. Unlike open-ended chat, we inject structural constraints into the instruction, limiting the generation search space to valid physical transitions.
\end{enumerate}

By wrapping the raw sequences in this rigorous format, we align the stochastic nature of physical dynamics with the deterministic syntax required for language modeling.

\subsection{Heterogeneous Task Integration}
\label{MultiTask}

To synergize domain-specific dynamic modeling with general scientific reasoning, we construct a heterogeneous instruction dataset comprising two distinct streams as shown in Figure \ref{fig2}(c).

\paragraph{Structured Forecasting Stream.}
We curate prediction tasks covering 1-step, 2-step, and backward trajectory forecasting. Crucially, we restrict the training horizon to short-term contexts (maximum 2 steps). By mastering local transition rules, the model is forced to acquire temporal inductive reasoning capabilities, enabling it to generalize to long-horizon (N-step) planning during inference without explicit supervision on long sequences.

\paragraph{Linguistic Regularization Stream.}

 While structured forecasting teaches the model the 'syntax' of reaction rules, it risks reducing chemical formulas to arbitrary symbols. To prevent catastrophic forgetting of general capabilities and provide semantic anchoring for chemical tokens, we interleave synthetic Chemistry Q\&A pairs. This stream serves as a semantic anchor, forcing the model to ground the symbolic molecular tokens in its pre-trained scientific knowledge base. This ensures that EvoMD-LLM evolves into a dual-capable agent: structurally grounded in specific physical dynamics while linguistically aligned with general chemical principles.

\subsection{Model Training and Inference}
As illustrated in Figure~\ref{fig2}(d), we employ a supervised fine-tuning (SFT) framework~\cite{ouyang2022training,taori2023stanford,unsloth,touvron2023llama}, aligning the model to predict target molecular events directly from structured input sequences. SFT enables EvoMD-LLM to internalize domain-specific transition rules into its parameters. The training process is explicitly designed to balance structural precision with linguistic generalization.

\paragraph{Input Representation and Architecture}
We utilize the Llama 3.1 8B \cite{meta2024llama3} backbone without architectural modifications. Formulas are tokenized using the standard Byte-Pair Encoding (BPE) \cite{sennrich2016bpe} vocabulary. This formatting encourages the model to process chemical formulas as semantic units, leveraging pretrained linguistic priors to model statistical regularities in molecular evolution.

\paragraph{Parameter-Efficient Optimization}
To align the model with reaction dynamics while preserving general scientific reasoning, we employ Low-Rank Adaptation (LoRA) \cite{hu2021lora}. By injecting trainable low-rank matrices into the attention and feed-forward layers while freezing pretrained weights, we achieve two strategic objectives: (1) prevention of catastrophic forgetting, ensuring the retention of the base model's linguistic priors essential for the QA component; and (2) training efficiency, which allows for rapid convergence on consumer-grade hardware by focusing capacity exclusively on modeling temporal chemical patterns.

\paragraph{Optimization Strategy}
To enable dual competence in structured forecasting and general reasoning, we employ a multi-task sampling strategy: structured prediction datasets and chemistry Q\&A instructions (Section~\ref{MultiTask}) are interleaved during training. The final loss is computed as the standard autoregressive cross-entropy over the target tokens of both tasks, ensuring the model simultaneously optimizes for domain-specific dynamics and linguistic fluency.

\section{Experiments}
We evaluate EvoMD-LLM on a range of temporal prediction tasks derived from molecular dynamics trajectories to assess its ability to model symbolic chemical evolution.

\subsection{Tasks and Evaluation Protocol}
To rigorously assess symbolic dynamic modeling, we utilize the Mo-S reactive system as a testbed. This system serves as a challenging benchmark due to its intrinsic stochasticity and the coexistence of competing reaction pathways (e.g., simultaneous growth and etching), demanding reasoning capabilities beyond simple pattern matching. This section evaluates whether EvoMD-LLM effectively addresses the proposed symbolic–temporal abstraction gap in modeling dynamic chemical systems. Specifically, we evaluate EvoMD-LLM on four temporal prediction tasks: 1-step prediction for short-range consistency, N-step prediction for iterative long-horizon forecasting, backward prediction for bidirectional reasoning of precursor states, and Potential-k prediction to capture the stochastic nature of chemical evolution \cite{coley2019graphreactivity}.

Performance is quantified using complementary metrics. Prediction accuracy and Potential-k accuracy track the presence of ground-truth states within the 1 and k predictions, respectively. These serve as proxies for logical consistency, assessing whether the model captures the valid causal logic of chemical evolution. Conversely, missing rate calculates the proportion of generated outputs that fail to parse as valid molecular formulas, thereby measuring the model’s Syntactic Validity and adherence to the chemical grammar. These metrics jointly assess the model's ability to navigate branching chemical reaction pathways while maintaining both structural integrity and instructional adherence.

\subsection{Experimental Setup}
\paragraph{Baseline Methods}
To evaluate the effectiveness of EvoMD-LLM, we compare it with four representative categories of baselines for temporal knowledge integration:
\begin{itemize}
    \item \textbf{Domain-specific LLM (ChemDFM)}: We include ChemDFM as a specialized chemistry-oriented LLM baseline because it is pretrained on chemical corpora and therefore provides a stronger domain-aware reference point than a general-purpose foundation model alone. We evaluate ChemDFM in both \textbf{zero-shot (ZS)} and \textbf{few-shot (FS)} settings using the same task instructions as EvoMD-LLM, where zero-shot receives only the test query and few-shot is provided with $k=3$ in-context trajectory examples.
    \item \textbf{In-context learning (ICL)} \cite{dong2024survey, luo2025promptchem}: We assess standard prompting capabilities including \textbf{ZS} ($k=0$) and \textbf{FS} ($k=3$) \cite{brown2020gpt3}. To probe scalability, we also implement \textbf{many-shot} (1,000 examples) \cite{agarwal2024many} and \textbf{full-Context} (7,321 examples) to test the upper bound of long-context reasoning.
    \item \textbf{Retrieval-Augmented generation (RAG)} \cite{lewis2020rag, zhong2025ragchem}: A dynamic memory baseline where a retriever selects the k most similar historical sub-sequences from the training set to provide input-dependent context.
    \item \textbf{Sequential baselines (Numerical Modality)}: To assess the necessity of symbolic abstraction over raw numerical fitting,  we consider two representative neural baselines that operate on numerical composition vectors (encoding atomic counts) rather than semantic tokens. Baselines include an LSTM \cite{hochreiter1997lstm} and a custom encoder-only Transformer \cite{vaswani2017attention}, which map trajectories to latent vectors for direct numerical regression.
\end{itemize}

\paragraph{Implementation Details}
Our final instruction-tuning dataset comprises over 22,766 samples, combining 7,321 stratified trajectory sequences with auxiliary scientific Q\&A data. For the generated RMD symbolic dataset, we adopt a trajectory-disjoint split: train and test examples are constructed from non-overlapping underlying MD trajectories, so no trajectory fragment, derived sub-sequence, or near-duplicate temporal context is shared across the two partitions. This protocol prevents leakage from the same simulated reaction path and provides a stricter evaluation of generalization to unseen trajectories. EvoMD-LLM is initialized with Llama 3.1 8B. We employ LoRA for parameter-efficient fine-tuning, optimizing approximately 42~M parameters ($r=16, \alpha=16$). The model is trained for 2 epochs using the AdamW optimizer \cite{loshchilov2019decoupled} with a global batch size of 8 and a peak learning rate of 2e-4. We utilize a linear learning rate scheduler and mixed-precision (bfloat16) with a maximum sequence length of 2048. All experiments are conducted on a single consumer-grade GPU (NVIDIA RTX 4090D) to demonstrate accessibility.

\begin{figure*}[tbp]
\centering
\includegraphics[width=0.95\textwidth]{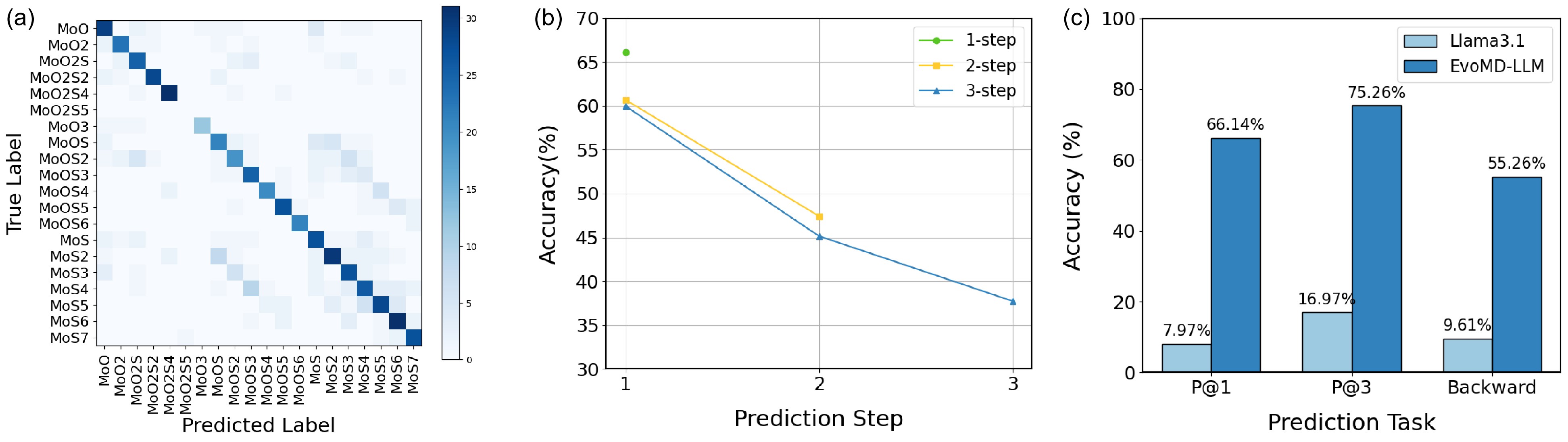}
\caption{Experimental results. (a) Confusion matrix showing discriminative capability. (b) Accuracy decay over N-step forecasting horizons. (c) Performance comparison against the LLaMA 3.1 base model across three tasks.}
\label{fig4}
\end{figure*}

\subsection{Overall Performance Comparison}
As shown in Table~\ref{tab:performance_comparison}, under the trajectory-disjoint split EvoMD-LLM significantly outperforms all baselines, achieving $66.14$\% accuracy with a 0\% missing rate across 10 runs. In comparison, the strongest retrieval-based baseline, RAG, reaches only $39.52$\% accuracy, while zero-shot Llama-3.1 exhibits a substantially higher missing rate of $36.50$\%. Notably, simply extending the context window (Content-1000/All) yields only marginal gains over few-shot prompting, with accuracy remaining below 20\%. This saturation suggests that naive long-context prompting fails to capture the temporal dependencies required for reaction forecasting. By contrast, supervised fine-tuning on symbolic trajectories enables EvoMD-LLM to learn structured temporal correlations directly, leading to substantially stronger predictive accuracy and syntactic stability. Statistical testing further confirms that the gain of EvoMD-LLM over the encoder-only sequential baseline is significant under a paired $t$-test ($p=0.01$), while its improvements over RAG and LSTM remain significant under Welch's $t$-test ($p<0.001$).

Figure~\ref{fig4}(a) presents the confusion matrix of molecular species predicted by the EvoMD-LLM. The strong diagonal dominance indicates robust discriminative capability, with only minor confusion among chemically similar or temporally adjacent species. This highlights the model’s ability to capture fine-grained symbolic and temporal distinctions within the molecular event space.

\subsection{Multi-step, Backward and Potential-k Prediction Analysis}

We further evaluate EvoMD-LLM on N-step, backward, and potential-k prediction tasks to assess its ability to model long-range temporal dependencies and stochastic reaction dynamics.

As shown in Figure~\ref{fig4}(b), accuracy decreases monotonically as the prediction horizon increases under the trajectory-disjoint split. Consistent with Table~\ref{tab:seq_comparison}, EvoMD-LLM achieves $66.14$\% for 1-step prediction, $53.21$\% for 2-step prediction, and $39.57$\% for 3-step prediction. This trend reflects error accumulation in autoregressive forecasting, while the remaining performance indicates that fine-tuning preserves substantial temporal consistency over multiple steps.

Figure~\ref{fig4}(c) shows that EvoMD-LLM substantially outperforms the LLaMA~3.1 base model in both potential-k and backward prediction. Under the same trajectory-disjoint evaluation, potential-1 performance is aligned with the 1-step result in Table~\ref{tab:performance_comparison}, and backward prediction reaches $55.26$\%, demonstrating the model's ability to infer plausible precursors from downstream states despite the stricter split.

\begin{figure*}[tbp]
\centering
\includegraphics[width=0.95\textwidth]{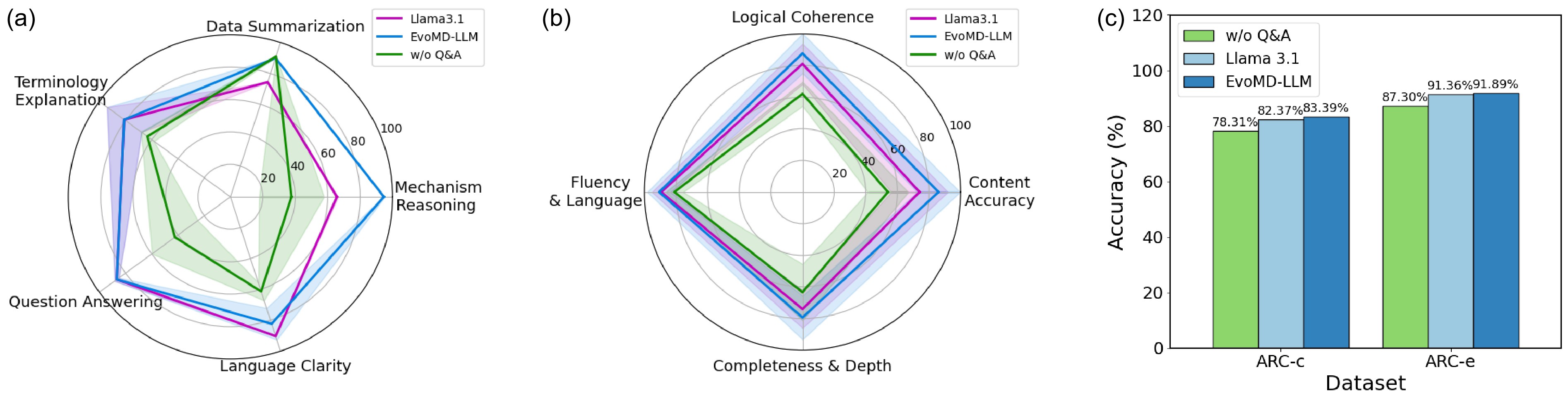}
\caption{Evaluation of general language understanding.
(a--b) present scores from an automated evaluation using Qwen3 as a judge.
(a) displays performance across five distinct capability dimensions, while (b) assesses four key criteria for output quality.
The shaded areas in the radar charts represent the standard deviation across the evaluation questions.
(c) shows the accuracy of the models on the ARC-c (Challenge) and ARC-e (Easy) benchmark datasets.}
\label{fig5}
\end{figure*}


\begin{table}[t]
\centering
\small
\begin{tabular}{lcc}
\hline
\textbf{Methods} & \textbf{Accuracy $\uparrow$} & \textbf{Missing Rate $\downarrow$} \\
\hline
\textit{Baselines (ChemDFM)} & & \\
Zero-shot  & $12.24 \pm 0.22$ & $1.36 \pm 0.17$ \\
Few-shot   & $8.60 \pm 0.92$  & $6.14 \pm 0.54$ \\
\hline
\textit{Baselines (Llama-3.1)} & & \\
Zero-shot  & $7.97 \pm 0.89$  & $36.50 \pm 2.40$ \\
Few-shot   & $16.94 \pm 0.76$ & $1.32 \pm 0.07$ \\
Content-1000 & $17.02 \pm 0.44$ & $1.09 \pm 0.06$ \\
Content-All  & $19.64 \pm 0.61$ & $0.08 \pm 0.12$ \\
RAG ($k=5$)  & $39.52 \pm 1.12$ & $0.92 \pm 0.28$ \\
\hline
\textit{Ours} & & \\
\textbf{EvoMD-LLM} & \textbf{66.14 $\pm$ 0.55} & \textbf{0.00} \\
\hline
\end{tabular}
\caption{Comparison of accuracy and missing rate. Methods are grouped by backbone models to eliminate redundancy and highlight the performance of EvoMD-LLM.}
\label{tab:performance_comparison}
\end{table}

\begin{table*}[t]
\centering
\small
\begin{tabular}{lcccc}
\hline
\textbf{Model} & \textbf{1-step} & \textbf{2-step} & \textbf{3-step} & \textbf{Backward} \\
\hline
LSTM & $38.35 \pm 1.42$ & $31.76 \pm 1.44$ & $26.19 \pm 0.71$ & $35.89 \pm 1.52$ \\
Encoder-only & $62.16 \pm 1.28$ & $41.84 \pm 1.63$ & $35.85\pm 1.93$ & $48.61 \pm 1.78$ \\
\textbf{EvoMD-LLM} & \textbf{66.14 $\pm$ 0.55} & \textbf{53.21 $\pm$ 0.47} & \textbf{39.57 $\pm$ 0.67} & \textbf{55.26 $\pm$ 1.25} \\
\hline
\end{tabular}
\caption{Performance comparison on molecular forecasting tasks. Reported values for N-step tasks represent the \textbf{average accuracy} over the entire prediction horizon (1 to N). Best results are \textbf{bolded}.}
\label{tab:seq_comparison}
\end{table*}

\subsection{Comparison with Sequential Baseline}

Table~\ref{tab:seq_comparison} compares EvoMD-LLM with an LSTM and an encoder-only model. For the 1-step prediction task, EvoMD-LLM achieves the highest accuracy at 66.14\%, outperforming both LSTM (38.35\%) and the encoder-only baseline (62.16\%).

As the prediction horizon increases, all methods exhibit performance degradation. EvoMD-LLM consistently maintains higher accuracy in the 2-step and 3-step settings, with particularly clear gains in the 3-step task. In the backward prediction task, EvoMD-LLM again outperforms both baselines. Overall, these results indicate that EvoMD-LLM provides more robust temporal modeling across different prediction settings.

\subsection{Ablation Studies and Analysis}
\label{sec:ablation}

\begin{table}[t]
\centering
\small
\begin{tabular}{lcc}
\hline
\textbf{Model Variant} & \textbf{1-Step} & \textbf{Backward} \\
\hline
w/o Temporal & 54.47 & 49.60 \\
w/o Q\&A & 63.38 & 53.03 \\
\textbf{EvoMD-LLM (Full)} & \textbf{66.14} & \textbf{55.26} \\
\hline
\end{tabular}

\caption{Ablation study. The significant drop in \textit{w/o Temporal} validates the necessity of kinetic scaffolding, while excluding Q\&A (\textit{w/o Q\&A}) has a relatively minor impact on domain-specific forecasting.}
\label{tab:ablation_temporal}
\end{table}

\definecolor{mechColor}{RGB}{178, 34, 34}   
\definecolor{stabColor}{RGB}{0, 102, 204}   
\newcommand{\hlMech}[1]{\textcolor{mechColor}{\textbf{#1}}}
\newcommand{\hlStab}[1]{\textcolor{stabColor}{\textbf{#1}}}


\begin{table*}[t]
  \centering
  \small
  \renewcommand{\arraystretch}{1.4}
  \begin{tabular}{p{0.2\linewidth} p{0.12\linewidth} p{0.62\linewidth}}
    \hline
    \textbf{Input History Sequence} & \textbf{Prediction} & \textbf{Generated Reasoning} \\
    \hline

    \multicolumn{3}{l}{\textbf{Case 1:  Trend Following (Oxygen $\to$ Sulfur Exchange)}} \\
    \texttt{(MoO,98); (MoOS$_2$,3); \newline (MoS,182)} &
    \textbf{Mol:} MoS$_3$ \newline \textbf{Time:} 106 ps &
    The transition from MoS to MoS$_3$ is driven by \hlMech{surface diffusion and adsorption} of additional MoS units. The predicted duration suggests the formation of a \hlStab{stable three-dimensional structure} favored by thermodynamics. \\

    \hline

    \multicolumn{3}{l}{\textbf{Case 2: Stability Detection (Distinguishing Stable vs. Transient)}} \\
    \texttt{(MoO$_2$,2); (MoO,29); \newline (MoOS$_2$,26)} &
    \textbf{Mol:} MoS \newline \textbf{Time:} 277 ps &
    The transition is driven by the \hlMech{decomposition of MoOS$_2$} due to high temperatures. The predicted duration (277 ps) suggests that MoS is \hlStab{relatively stable}, forming a \hlStab{metastable trap} at these conditions. \\

    \hline

    \multicolumn{3}{l}{\textbf{Case 3: Structure Building (Increasing Complexity)}} \\
    \texttt{(MoOS$_2$,43); (MoOS$_4$,6); \newline (MoS$_3$,7)} &
    \textbf{Mol:} MoS$_5$ \newline \textbf{Time:} 365 ps &
    The transition involves the \hlMech{addition of two sulfur atoms} through a series of elementary reactions. This process reflects the increasing preference for \hlStab{higher-order coordination structures} in Mo-S systems. \\

    \hline
  \end{tabular}
  \caption{\label{tab:qualitative_cases}
    Qualitative examples of reasoning generated by EvoMD-LLM.
    The selected cases demonstrate the model's ability to track linear evolutionary pathways without oscillation.Highlights indicate the textual description of \textbf{reaction mechanisms (Red)} and \textbf{stability/metastability (Blue) }aligned with temporal cues.
  }
\end{table*}

\subsubsection{Impact of Temporal Scaffolding}
As detailed in Table \ref{tab:ablation_temporal}, excising the duration targets leads to a consistent performance degradation. The 1-step prediction accuracy declines from 66.14\% to 54.47\%, with a comparable drop in backward reasoning. These results empirically validate that temporal supervision is not merely an auxiliary task but a necessary constraint for correct chemical reasoning.

\subsubsection{Impact of Multi-Task Instruction Tuning}

\label{sec:language_eval}
A key design goal of EvoMD-LLM is to improve structured molecular forecasting without degrading general scientific language capabilities. To assess this, we evaluate an ablated variant trained without the chemistry Q\&A dataset (w/o Q\&A).

We adopt both qualitative and quantitative evaluations. For qualitative assessment, we use Qwen3 \cite{yang2025qwen3} as an automated evaluator to score model responses across multiple scientific capability and quality dimensions. In addition, we report performance on the AI2 Reasoning Challenge (ARC) benchmark \cite{clark2018arc} to measure standardized scientific reasoning, containing a total of 3548 test samples.

As shown in Figure~\ref{fig5}(a–b), removing Q\&A supervision leads to consistent degradation across all evaluated dimensions, with the largest drops observed in mechanism reasoning and question answering. The ablated model also exhibits reduced coherence and fluency, suggesting that natural language supervision contributes to stable scientific expression.

Figure~\ref{fig5}(c) further shows that EvoMD-LLM maintains strong performance on both ARC-e and ARC-c, comparable to the base LLaMA~3.1 model, while the w/o Q\&A variant performs noticeably worse. These results indicate that the Q\&A dataset plays an important role in preserving general scientific reasoning during domain-specific fine-tuning.

\subsection{Qualitative Analysis}
\label{sec:qualitative_analysis}

Quantitative metrics summarize predictive accuracy but provide limited insight into model behavior. We therefore present a qualitative analysis to examine the how EvoMD-LLM explains its predictions without explicit supervision on reaction mechanisms. This analysis focuses on the alignment of learned sequence patterns with semantic explanations, rather than on validating ab initio physical correctness.

Table~\ref{tab:qualitative_cases} demonstrates the model's context awareness. For instance, it correctly links early-stage sulfidation to surface diffusion (Case~1) while identifying high-temperature decomposition in intermediate phases (Case~2). Furthermore, it utilizes the duration token as a semantic pivot to differentiate between kinetically stable products and metastable traps, as evidenced by the distinct duration predictions. Additional qualitative examples, including typical failure modes, are provided in Appendix~\ref{app:qualitative_examples}.

\section{Conclusion}

In this work, we introduce EvoMD-LLM, a framework that re-frames molecular dynamics as a symbolic language modeling problem, thereby internalizing the "grammar" of chemical evolution into LLMs. By aligning continuous physical trajectories with discrete semantic tokens through an Temporal Scaffolding strategy, we enable the model to treat temporal persistence as a semantic component. We demonstrate that this design introduces a robust inductive bias toward temporally consistent generation, leading to improved forecasting accuracy and a substantial reduction in invalid molecular states. More broadly, EvoMD-LLM highlights the potential of language-based models as general-purpose sequence learners for scientific simulations, suggesting a promising direction for bridging linguistic abstraction with time-resolved molecular dynamics in AI-driven materials discovery.

\section*{Limitations}

While EvoMD-LLM demonstrates promising capabilities in modeling symbolic chemical evolution, several limitations remain to be addressed in future work:

\textbf{Generalization to Unseen Chemical Spaces.} Our evaluation focuses on the Mo-S CVD system. We selected this system not merely for data availability, but as a representative "complex prototype" of inorganic synthesis: it features high-degree stochastic branching, reversibility, and multi-phase transitions (nucleation, etching, growth), which are often absent in linear organic reaction datasets. However, extending this framework to diverse chemical spaces, including heterogeneous biological systems, remains an open challenge for future scaling.

\paragraph{Autoregressive Error Accumulation.}
As observed in N-step prediction tasks, the model suffers from error accumulation typical of autoregressive generation, leading to performance degradation over long horizons. Unlike numerical solvers that strictly enforce conservation laws, the current probabilistic generation may occasionally drift into physically invalid states. Integrating physical constraints (e.g., mass conservation or energy consistency) directly into the loss function could mitigate this issue in future iterations.

\paragraph{Loss of Fine-Grained Geometry.}
While our coarse-grained symbolic representation efficiently filters thermal noise and captures high-level reaction logic, it inevitably discards fine-grained conformational information (e.g., precise bond lengths and angles). Consequently, EvoMD-LLM is currently less suitable for tasks requiring exact geometric verification. Future work could explore multi-modal architectures that jointly model symbolic evolution and geometric deformation to achieve fully comprehensive dynamic reasoning.

\paragraph{Interpretability and Hallucination Risks.}
 The explanatory outputs provided by EvoMD-LLM are derived from aligning trajectory patterns with scientific knowledge learned during training, rather than ab initio derivation. Consequently, explanations often rely on plausible but geometrically ungrounded terminology, suggesting retrieval-based association driven by pre-trained priors. Furthermore, the model occasionally over-interprets stochastic cues as deterministic stability guarantees and defaults to generic linear narratives for rare intermediates, reflecting reduced sensitivity in low-frequency regimes.

\section*{Acknowledgement}
The authors would like to thank the support from the Science and Technology Commission of Shanghai Municipality (No. 25DZ3001902). This work was partially supported by SJTU Kunpeng\&Ascend Center of Excellence.

\bibliography{custom}

\appendix



\section{Data Processing and Statistics}
\label{app:data_stats}

\begin{figure*}[htbp]
\centering
\includegraphics[width=1\linewidth]{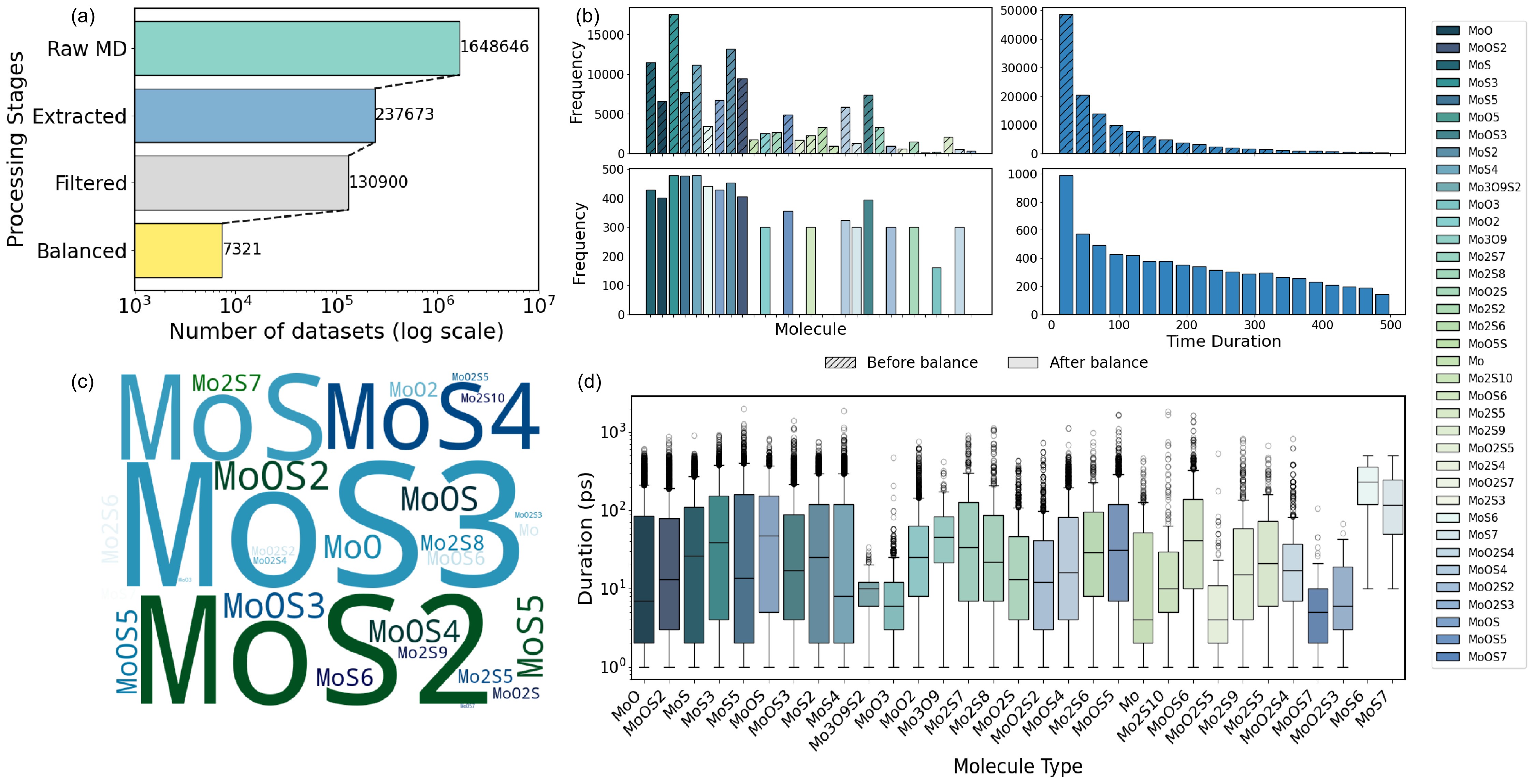}
\caption{Data processing visualization. (a) Evolution of dataset scale across successive preprocessing stages, showing the reduction from raw events to high-quality balanced sequences. (b) Histograms comparing molecular type and duration distributions before (Origin) and after (Processed) stratified sampling, highlighting the mitigation of the long-tail problem. (c) Word cloud visualizing the dominance of specific species in the raw dataset. (d) Box plots showing the distinct distribution of existence durations (in ps) for different molecular species, reflecting their varying kinetic stabilities.}
\label{fig3}
\end{figure*}

In this section, we provide a detailed breakdown of the data processing pipeline, statistical characteristics, and the balancing strategy visualized in Figure \ref{fig3}.
\subsection{Event Extraction and Filtering Pipeline}
Our molecular event sequences originate from Reactive Molecular Dynamics (RMD) simulations of MoS$_2$ synthesis, as reported by \newcite{dang2025unraveling}. The raw trajectories capture high-frequency atomic motions that do not directly correspond to symbolic chemical reactions. To align this data with language modeling, we employed a multi-stage pipeline:

\begin{itemize}
    \item \textbf{Raw Extraction (Step 1):} We converted each MD frame into an undirected graph $G=(V,E)$ using bond-order cutoffs, with atoms as nodes and valid bonds as edges. A Depth-First Search (DFS) traversal was then used to identify connected components, each of which was serialized into a molecular formula. This yielded an initial "Raw MD" dataset comprising 1,648,646 events (Figure \ref{fig3}(a)).
    \item \textbf{Thermal Noise Reduction (Step 2):} Raw trajectories are dominated by transient thermal fluctuations where bonds vibrate but do not break. We therefore discard all events with $\Delta t < \tau_{\min}$ as high-frequency noise, with $\tau_{\min}=10$ ps. This resulted in the "Extracted" dataset of 237,673 events.
    \item \textbf{Temporal Band-Pass Filtering (Step 3):} To focus on the primary dynamic scales relevant for reaction forecasting, we further refined the dataset by retaining only events satisfying $\tau_{\min} \leq \Delta t \leq \tau_{\max}$, where $(\tau_{\min}, \tau_{\max}) = (10, 500)$ ps. This step removes extremely short-lived noise while excluding ultra-long plateau states, resulting in the "Filtered" dataset of 130,900 events.

\end{itemize}

\subsection{Handling Data Imbalance}

A critical challenge in MD data is the long-tail distribution of species. As shown in the "Origin Data" histograms in Figure \ref{fig3}(b) and the visual representation in the word cloud (Figure \ref{fig3}(c)), a few dominant species (e.g., reactants like MoS$_x$ precursors) account for the vast majority of observations, while critical transition states are rare.


Training a language model directly on this skewed distribution leads to trivial solutions where the model simply memorizes the most frequent tokens. To address this, we applied stratified sampling to balance the dataset across both molecular identity and duration intervals.
\begin{itemize}
    \item \textbf{Effect of Balancing:} Figure \ref{fig3}(b) (bottom panels) demonstrates that after sampling, the distributions of both molecular types and event durations become significantly more uniform.
    \item \textbf{Final Dataset:} This process yielded the final "Balanced" dataset containing 7,321 high-quality sequence pairs, which were used for fine-tuning EvoMD-LLM.
\end{itemize}

\subsection{Temporal Characteristics}
Figure \ref{fig3}(d) presents box plots of the existence durations for various molecular species in the final processed dataset. The distinct temporal distributions (e.g., some species consistently show shorter lifetimes than others) confirm that duration is a semantic property intrinsic to each chemical species, justifying our use of Temporal Scaffolding to capture these kinetic signatures.

\section{Implementation Details}
\label{sec:appendix_implementation}

To ensure the reproducibility of EvoMD-LLM, we provide detailed specifications of our software environment, hardware infrastructure, and training configurations.

\subsection{Software and Hardware Environment}
We implemented EvoMD-LLM using the \texttt{Unsloth} framework, which optimizes memory usage and training speed for Llama-based models. The core software dependencies include:
\begin{itemize}
    \item \textbf{Python}: 3.10
    \item \textbf{PyTorch}: 2.7.0 (with CUDA 12.6)
    \item \textbf{Unsloth}: 2025.5.6
    \item \textbf{Transformers}: 4.51.3
    \item \textbf{Peft}: 0.15.2
\end{itemize}

All experiments were conducted on a single consumer-grade NVIDIA RTX 4090D (24GB VRAM) GPU. We utilized mixed-precision training (\texttt{bfloat16}) to maximize computational efficiency without compromising numerical stability.

\subsection{Details on Quantitative Evaluation}
All reported quantitative results are computed as the mean over multiple runs with different random seeds to ensure robust and reproducible evaluation. Specifically, for each experiment, we performed 10 runs with distinct random seeds and report the averaged metrics.

For the additional reasoning assessment on the ARC benchmark, we utilized the OpenCompass evaluation platform \cite{2023opencompass}. The model was evaluated using the \texttt{HuggingFaceCausalLM}.

\subsection{Hyperparameters and Training Costs}
The detailed hyperparameters used for fine-tuning are listed in Table~\ref{tab:s1_hyperparams}, optimized by grid search. We employed the LoRA technique, targeting all linear layers in the attention and feed-forward blocks.

With the configuration specified below, the full training process (covering both structured forecasting and Q\&A tasks) took approximately 2.5 hours. The peak memory usage was controlled under 16GB thanks to the 4-bit quantization support from Unsloth during the gradient calculation.

\begin{table}[h]
    \centering
    \small
    \caption{Detailed hyperparameters and training configuration for EvoMD-LLM.
        The model was fine-tuned using the LoRA method with the Unsloth framework for memory optimization.}
    \label{tab:s1_hyperparams}
    \begin{tabular}{ll}
        \hline
        \textbf{Hyperparameter} & \textbf{Value} \\
        \hline
        \textit{General Configuration} & \\
        Base Model & Llama 3.1 8B Instruct \\
        Framework & Unsloth (TRL) \\
        Precision & bfloat16 (bf16) \\
        Random Seed & 3407 \\
        Max Sequence Length & 2048 \\
        \hline
        \textit{Optimization} & \\
        Optimizer & AdamW (8-bit) \\
        Learning Rate & $2 \times 10^{-4}$ \\
        Weight Decay & 0.01 \\
        LR Scheduler & Linear \\
        Warmup Steps & 400 \\
        Num Epochs & 2 \\ 
        \hline
        \textit{Batch Size Configuration} & \\
        Per-Device Batch Size & 2 \\
        Gradient Accumulation Steps & 4 \\
        Effective Batch Size & 8 \\
        \hline
        \textit{LoRA Configuration} & \\
        Rank ($r$) & 16 \\
        Alpha ($\alpha$) & 16 \\
        Dropout & 0 \\
        Bias & None \\
        Target Modules& q, k, v, o, gate, \\
                       & up, down\_proj \\
        \hline
    \end{tabular}
\end{table}

\subsection{Licenses and Terms of Use.}
We use the LLaMA 3.1 model released by Meta under the LLaMA community license.
The ARC-e and ARC-c benchmarks are publicly available for research purposes.
All reactive molecular dynamics simulations and derived symbolic datasets were
generated by the authors and do not contain personal or sensitive information.
We plan to release the processed datasets and model checkpoints under a
permissive research license upon acceptance.

\section{Prompt Templates}
\label{sec:appendix_prompts}

In this section, we present the exact prompt templates used for training EvoMD-LLM and for eliciting qualitative reasoning. We employed a consistent system prompt to define the model's role, while task-specific instructions were appended to the user queries to constrain the output format.

\subsection{Training and Prediction Prompts}
For the supervised fine-tuning (SFT) stage and standard prediction tasks (1-step, N-step, and Backward), we used the following template structure.

\paragraph{System Message.} This prompt sets the general behavioral constraints and defines the data format (molecule, duration).
\begin{quote}
\small\ttfamily
You are an AI assistant to help me predict molecular sequence progression based on given molecular compositions and their existence durations and analysis. Each data point consists of a molecule and the duration it persists in the system, the unit of duration is ps. If the question is about predicting molecular sequences, format your answer as (molecule, time). Otherwise, answer normally.
\end{quote}

\paragraph{Task-Specific Instructions.} Different prediction tasks are distinguished by specific suffixes appended to the historical sequence.

\noindent\textbf{1. Single-Step Prediction (Forward):}
\begin{quote}
\small\ttfamily
\textbf{Input:} The history sequence is \{SEQUENCE\_HISTORY\}, What is the next element? Output ONLY the next element in the format: (molecule, time). No explanation. No code. No extra words!
\end{quote}

\noindent\textbf{2. Multi-Step Prediction (N=2):}
\begin{quote}
\small\ttfamily
\textbf{Input:} The history sequence is \{SEQUENCE\_HISTORY\}, What are the next two elements? Output ONLY the next two elements in the format: (molecule, time). No explanation. No code. No extra words!
\end{quote}

\noindent\textbf{3. Backward Prediction:}
\begin{quote}
\small\ttfamily
\textbf{Input:} The history sequence is \{SEQUENCE\_HISTORY\}, What is the previous element? Output ONLY the previous element in the format: (molecule, time). No explanation. No code. No extra words!
\end{quote}

\section{Reasoning and Explanation Prompts}
\label{sec:appendix_prompts}

To assess the emergent explanatory capabilities of EvoMD-LLM, we utilized a structured prompt designed to constrain the output format.

It is important to note that while the model was fine-tuned with a mixture of symbolic MD sequences and general scientific Q\&A pairs (to prevent catastrophic forgetting, see Section 2.5), it was never supervised on paired samples of (trajectory, textual explanation).

The training data for MD trajectories consisted solely of symbolic sequences (e.g., molecule tokens and duration values). Therefore, the detailed reasoning elicited by the prompt below reflects the model's emergent ability to ground its general chemical knowledge (acquired from pre-training and Q\&A regularization) into the specific context of the learned physical dynamics.

The prompt acts as a structural scaffold, directing the model to articulate its learned sequence patterns into explicit linguistic reasoning.


\paragraph{Expert Simulator System Context.}
\begin{quote}
\small\ttfamily
You are an expert scientific simulator specializing in Reactive Molecular Dynamics (RMD) for Chemical Vapor Deposition (CVD) synthesis.

\textbf{System Context:}
The reaction system involves the sulfidation of Mo3O9 precursors by S2 gas. Key dynamics include Oxygen-Sulfur exchange, structural relaxation, and thermal decomposition.

\textbf{Task Definition:}
Your goal is to forecast the trajectory of chemical evolution. Each data point (Molecule, Duration) represents a distinct chemical state and its kinetic persistence (stability).
\begin{itemize}
    \item A short duration implies a transient intermediate or transition state.
    \item A long duration implies a thermodynamically stable product or metastable trap.
\end{itemize}
\end{quote}

\paragraph{Reasoning Instruction.} After the model generates a prediction, we prompt it to explain the physical rationale using the following template:

\begin{quote}
\small\ttfamily
\textbf{Task:} You are provided with a historical trajectory of molecular species and their durations.

\textbf{History Sequence:} \{history\_seq\} \\
\textbf{Your Model Prediction:} (\{predict\_res\})

\textbf{Instructions:}
Provide a scientific explanation for this transition. Your response must:
\begin{enumerate}
    \item \textbf{Mechanism:} Analyze the change in stoichiometry from the last history step to the predicted step. What specific chemical process drives this transformation?
    \item \textbf{Stability:} Analyze the predicted duration (\{duration\}). What does this specific timescale imply about the thermodynamic state or kinetic stability of the predicted molecule?
    \item \textbf{Format:} Write in strict, concise Academic English.
\end{enumerate}
Your answer must be in academic English, concise, and only include the reasoning (no extra content, no repetition).
\end{quote}

\section{Sample Efficiency Analysis}
\label{app:sample_efficiency}

\begin{figure}[h]
    \centering
    \includegraphics[width=1\linewidth]{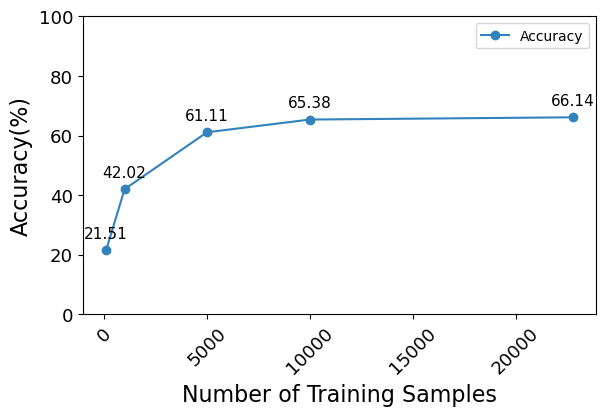}
    \caption{Learning curve of EvoMD-LLM. The plot shows 1-step prediction accuracy scaling with training data size. The model exhibits strong few-shot generalization, reaching over 60\% accuracy with only 5,000 samples, and shows signs of performance saturation beyond 10,000 samples, indicating that the current dataset size is sufficient for capturing the core dynamics.}
    \label{fig:learning_curve}
\end{figure}

To investigate whether our dataset size is a bottleneck for performance, we conducted a scaling analysis by training EvoMD-LLM on subsets of the training data ranging from roughly 100 to 22,000 samples. Figure~\ref{fig:learning_curve} illustrates the 1-step prediction accuracy as a function of data quantity.

As shown in Figure~\ref{fig:learning_curve}, the model exhibits high sample efficiency.
\begin{itemize}
    \item \textbf{Rapid Syntax Acquisition (0-5k):} Accuracy surges from 21.5\% to 61.1\% within the first 5,000 samples. This steep rise suggests that the LLM, leveraging its pre-trained capabilities, rapidly aligns with the "grammar" of molecular evolution (syntax and basic stoichiometry) with minimal data.
    \item \textbf{Performance Saturation (10k-20k):} As data volume doubles from 10,000 to 22,000, accuracy gains moderate (from 65.4\% to 66.1\%). This plateau indicates that the model has effectively captured the majority of the learnable patterns within the current domain.
\end{itemize}

This analysis confirms that our dataset size ($\sim$20k total samples) is robust. The constraint on further performance improvement is likely not the quantity of raw data, but the inherent stochasticity of the chemical system itself.

\section{Detailed Error Analysis}
\label{sec:appendix_error_analysis}

\begin{figure}[h]
    \centering
    \includegraphics[width=0.5\textwidth]{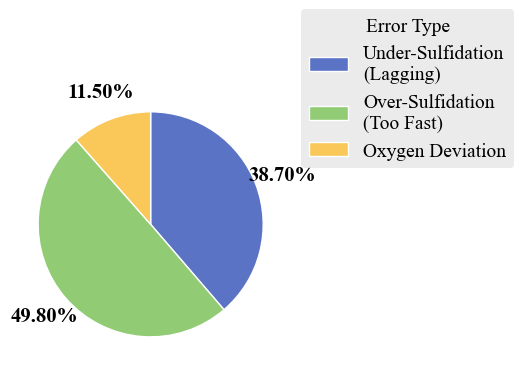}
    \caption{Breakdown of Kinetic Mismatch Errors. The distribution shows a balanced split between Under-Sulfidation (Blue, 38.7\%) and Over-Sulfidation (Green, 49.8\%). This symmetry indicates that the model's temporal errors are stochastic rather than biased. Oxygen Deviation (Yellow, 11.5\%) represents minor stoichiometric noise.}
    \label{fig:error_breakdown}
\end{figure}

To evaluate the reliability of EvoMD-LLM beyond standard accuracy metrics, we conducted a fine-grained analysis of the prediction errors on the test set.

\subsection{Zero Hallucinations and Chemical Validity}

A critical finding of our analysis is that EvoMD-LLM exhibits zero hallucinations regarding chemical validity. All of the incorrect predictions correspond to chemically valid molecular formulas that exist within the reaction network (e.g., predicting a valid intermediate like $MoS_3$ but at an incorrect time step).

This stands in sharp contrast to generic LLMs, which often generate physically impossible stoichiometry when fine-tuned on scientific data. The absence of hallucinations confirms that our symbolic tokenization strategy has successfully grounded the model in the compositional grammar of the chemical system, constraining its errors to the domain of physical kinetics rather than generative syntax.

\subsection{Physical Symmetry of Kinetic Mismatch}
Since all errors are valid "Kinetic Mismatches," we further decomposed them to determine if the model exhibits systematic bias. Given that the primary reaction mechanism is sulfidation (replacing Oxygen with Sulfur), we classified the errors based on the stoichiometry of the predicted species relative to the ground truth:

\begin{itemize}
    \item \textbf{Under-Sulfidation (Lagging):} The predicted molecule contains fewer Sulfur atoms than the ground truth ($S_{pred} < S_{true}$). The model predicts a precursor state, effectively "lagging" behind the true trajectory.
    \item \textbf{Over-Sulfidation (Too Fast):} The predicted molecule contains more Sulfur atoms ($S_{pred} > S_{true}$). The model anticipates the reaction progressing faster than reality.
    \item \textbf{Oxygen Deviation:} The Sulfur content is correct, but the Oxygen stoichiometry differs, reflecting minor inaccuracies in secondary deoxidation steps.
\end{itemize}


As illustrated in Figure~\ref{fig:error_breakdown}, the distribution of kinetic mismatch errors exhibits a broadly balanced profile, despite the inherent complexity of the Mo-S reactive system. While a slight divergence is observed where Over-Sulfidation (49.8\%) marginally exceeds Under-Sulfidation (38.7\%), the results suggest that EvoMD-LLM effectively avoids severe systematic drift towards either accelerating or retarding the reaction kinetics. The slight prevalence of over-sulfidation errors likely reflects the model sensitivity to high frequency sulfidation events that dominate the CVD growth phase. Overall, the distribution confirms that the errors primarily represent unbiased variance stemming from the stochastic nature of Molecular Dynamics simulations, where atomic transitions fluctuate around the mean reaction path. By capturing this central tendency, the framework demonstrates its capability to internalize the underlying grammar of chemical evolution without falling into deterministic traps.

\section{Additional Qualitative Examples}
\label{app:qualitative_examples}
To provide a more comprehensive view of the model’s qualitative behavior, we present additional prediction and reasoning examples in Table~\ref{tab:add_qualitative_cases}.Unlike the selected cases in the main paper, these examples include both successful predictions and characteristic failure modes, and are intended to illustrate typical patterns rather than exhaustive coverage.
\begin{table*}[t]
  \centering
  \small
  \renewcommand{\arraystretch}{1.35}
  \begin{tabular}{p{0.22\linewidth} p{0.12\linewidth} p{0.6\linewidth}}
    \hline
    \textbf{Input History Sequence} & \textbf{Prediction} & \textbf{Generated Reasoning} \\
    \hline

    \multicolumn{3}{l}{\textbf{Case 1: Over-sulfidation Drift}} \\
    \texttt{(MoS$_4$,8); (Mo$_2$S$_7$,12); \newline (MoS$_3$,41)} &
    \textbf{Mol:} MoS$_6$ \newline \textbf{Time:} 10 ps &
    The transition is attributed to \hlMech{continued sulfur adsorption} leading to higher sulfur coordination. However, the extremely short predicted duration suggests a \hlStab{transient and unstable configuration}. \\

    \hline
    \multicolumn{3}{l}{\textbf{Case 2: Stability Overestimation}} \\
    \texttt{(Mo$_2$S$_5$,7); (MoS$_3$,19); \newline (MoS$_4$,32)} &
    \textbf{Mol:} MoS$_4$ \newline \textbf{Time:} 125 ps &
    The model interprets repeated appearance of MoS$_4$ as evidence of \hlStab{kinetic stability}, potentially overestimating persistence due to frequency bias. \\

    \hline
    \multicolumn{3}{l}{\textbf{Case 3: Reduction Shortcut}} \\
    \texttt{(MoOS,4); (MoO$_2$,9); \newline (MoO$_2$,12)} &
    \textbf{Mol:} MoO \newline \textbf{Time:} 488 ps &
    The transition is explained as \hlMech{oxygen loss under reducing conditions}. The long duration reflects a \hlStab{stable oxide plateau}, though intermediate pathways are omitted. \\

    \hline
    \multicolumn{3}{l}{\textbf{Case 4: Missed Cluster Growth}} \\
    \texttt{(Mo$_2$S$_7$,5); (Mo$_2$S$_5$,11); \newline (MoS$_4$,18)} &
    \textbf{Mol:} MoS$_4$ \newline \textbf{Time:} 495 ps &
    The model favors \hlStab{structural persistence} over further aggregation, overlooking potential \hlMech{cluster growth} toward higher sulfur content. \\

    \hline
    \multicolumn{3}{l}{\textbf{Case 5: Correct Stability Tracking}} \\
    \texttt{(MoS$_4$,2); (MoOS$_4$,9); \newline (MoOS$_5$,14)} &
    \textbf{Mol:} MoOS$_5$ \newline \textbf{Time:} 62 ps &
    The prediction reflects \hlMech{incremental sulfur incorporation} while correctly identifying a \hlStab{moderately persistent intermediate}. \\

    \hline
    \multicolumn{3}{l}{\textbf{Case 6: Linear Growth Bias}} \\
    \texttt{(MoS$_3$,6); (MoS$_4$,11); \newline (MoS$_5$,9)} &
    \textbf{Mol:} MoS$_6$ \newline \textbf{Time:} 83 ps &
    The reasoning assumes monotonic sulfur accumulation, exhibiting a \hlMech{linear growth bias} that neglects possible sulfur loss or rearrangement. \\

    \hline
    \multicolumn{3}{l}{\textbf{Case 7: Transient Misclassification}} \\
    \texttt{(MoO$_2$,3); (MoOS$_2$,7); \newline (MoS$_2$,5)} &
    \textbf{Mol:} MoS$_2$ \newline \textbf{Time:} 214 ps &
    Although MoS$_2$ is predicted as long-lived, the explanation likely overstates \hlStab{kinetic trapping} for what may be a transient configuration. \\

    \hline
    \multicolumn{3}{l}{\textbf{Case 8: Rare Species Generalization}} \\
    \texttt{(Mo$_3$S$_{11}$,4); (Mo$_3$S$_{12}$,6); \newline (Mo$_3$S$_{13}$,3)} &
    \textbf{Mol:} Mo$_3$S$_{13}$ \newline \textbf{Time:} 57 ps &
    The model produces a generic explanation invoking \hlMech{coordination saturation}, reflecting limited specificity for \hlStab{rare cluster species}. \\

    \hline
    \multicolumn{3}{l}{\textbf{Case 9: Oxygen Retention Bias}} \\
    \texttt{(MoO$_3$,5); (MoO$_2$,18); \newline (MoOS$_2$,7)} &
    \textbf{Mol:} MoO$_2$ \newline \textbf{Time:} 301 ps &
    The prediction favors oxygen-rich species, suggesting a bias toward \hlStab{oxide persistence} despite emerging sulfidation signals. \\

    \hline
    \multicolumn{3}{l}{\textbf{Case 10: Competing Pathway Suppression}} \\
    \texttt{(MoS$_2$,14); (MoOS$_3$,9); \newline (MoS$_3$,11)} &
    \textbf{Mol:} MoS$_4$ \newline \textbf{Time:} 92 ps &
    The explanation emphasizes sulfur addition while suppressing alternative \hlMech{desulfurization or rearrangement pathways}. \\

    \hline
  \end{tabular}
  \caption{Additional qualitative examples generated by EvoMD-LLM.
These cases complement the main-paper examples by covering a broader range of behaviors,
including correct predictions, stability overestimation, linear growth bias, and generic reasoning for rare species.
Highlights indicate inferred reaction mechanisms (\hlMech{Red}) and stability or metastability judgments (\hlStab{Blue}).}
\label{tab:add_qualitative_cases}
\end{table*}

\end{document}